\documentclass[10pt, a4paper]{article}

\usepackage{lrec-coling2024}
\usepackage{CJKutf8}
\usepackage{amsmath}
\usepackage{listings}
\usepackage{color}
\usepackage{xcolor}
\usepackage{url}
\usepackage{textcomp}
\usepackage{pgfplots}
\usepackage{pdfpages}
\usepackage{adjustbox}
\usepackage{graphicx} 
\usepackage{multirow}
\usepackage{booktabs}
\DeclareMathOperator*{\argmax}{arg\,max}

\newcommand*{\red}{\textcolor{black}}

\title{An Automatic Quality Metric for Evaluating Simultaneous Interpretation}

\name{Mana Makinae$^1$, Katsuhito Sudoh$^1$, Masaru Yamada$^2$, Satoshi Nakamura$^1$} 

\address{$^1$Nara Institiute of Science and Technology, 8916-5 Takayama, Ikoma, Nara 630-0192, Japan \\
         $^2$Rikkyo University, 3-34-1 Nishi-Ikebukuro, Toshima-ku, Tokyo 171-8501, Japan \\
         \{makinae.mana.mh2, sudoh, s-nakamura\}@is.naist.jp\\}

\abstract{
Simultaneous interpretation (SI), the translation of one language to another in real time, starts translation before the original speech has finished.
Its evaluation needs to consider both latency and quality. This trade-off is challenging especially for distant word order language pairs such as English and Japanese.
To handle this word order gap, interpreters maintain the word order of the source language as much as possible to keep up with original language to minimize its latency while maintaining its quality, whereas in translation reordering happens to keep fluency in the target language.
This means outputs synchronized with the source language are desirable based on the real SI situation, and it's a key for further progress in computational SI and simultaneous machine translation (SiMT).
In this work, we propose an automatic evaluation metric for SI and SiMT focusing on word order synchronization.
Our evaluation metric is based on rank correlation coefficients, leveraging cross-lingual pre-trained language models.
Our experimental results on NAIST-SIC-Aligned and JNPC showed our metrics' effectiveness to measure word order synchronization between source and target language.
The code is available at \url{https://github.com/ahclab/Combined_Metrics}.
 \\ \newline  }

\begin{document}

\maketitleabstract

\section{Introduction}
Simultaneous Interpretation (SI) is the task of delivering translation in real-time, allowing the audience to understand the contents without significant delays.
One challenge is to balance the trade-off between latency and quality, which is particularly difficult
with different word-order language pairs such as English and Japanese.
Interpreters avoid word reordering and keep source language word order as much as possible to achieve good quality and minimal delay with limited working memory.
This is called the first-in-first-out (FIFO) strategy.

While there have been growing research interests in computational approaches to SI, called Simultaneous Machine Translation (SiMT), most previous studies \cite{gu-etal-2017-learning,ma-etal-2019-stacl} still rely largely on offline translation data due to the size limitation of SI corpora.
Such offline translation data are not sufficient to learn the FIFO strategy, so the SiMT still suffers from difficulty in translating in the FIFO manner.
Table~\ref{tab:simul_examples} shows an example of English-Japanese word order differences between offline translation and SI.
Consider the SI: in English, the words ``every year'' come at the end of the sentence, whereas in Japanese the translation of ``every year'' comes at the middle.
This indicates the interpretation has already started in the middle of the source speech to reduce the latency.
On the other hand, ``every year'' is translated at the beginning of the offline translation.
This indicates the translation needs to wait until the end of this source speech segment, which increases the latency. 

\begin{table*}
\centering
\begin{tabular}{cp{13cm}}
\toprule
Source & {\small (1) Out of seven / (2) large public corporations / (3) commit / (4) frauds / (5) \textbf{every year}.} \\
\midrule
Simultaneous & \begin{CJK}{UTF8}{min}{\small 上場している企業の / 7社に1社は /  \underline{毎年} / 不正行為を / しています。}\end{CJK} \\
Interpretation & {\small [ (2) large public corporation / (1) out of seven / (5) \textbf{every year} / (4) frauds / (3) commit ]}\\
\midrule
Offline & \begin{CJK}{UTF8}{min}{\small \underline{毎年} / 大企業の / 7社に1社が / 不正行為を / 働いています。}\end{CJK} \\
Translation & {\small [ (5) \textbf{every year} / (2) large public corporation / (1) out of seven / (4) frauds / (3) commit ]} \\
\bottomrule
\end{tabular}
\caption{Example of word order differences}
\label{tab:simul_examples}
\end{table*}

In this work, we aim to evaluate this word order difference in SI and SiMT to estimate their possible latency increase.
Word reordering negatively affects the performances for both human SI and computational SiMT.
For this reason, word order synchronization is essential to reduce latency and needs to be addressed as one of the aspects for its evaluation. 
We propose an evaluation metric to measure word order synchronization between source language inputs and target language outputs for SI and SiMT, inspired by previous work on automatic machine translation evaluation.
We approach to this problem leveraging the methodology of BERTScore \cite{Zhang2020BERTScore}
to find source-target word alignment for calculating word order synchronization as a rank correlation coefficient as \citet{isozaki-etal-2010-automatic}.
The vector similarity-based alignment in BERTScore can be easily extended to our cross-lingual problem using 
multilingual BERT \cite{devlin-etal-2019-bert}.
We investigate the usefulness of the proposed metric using an English-to-Japanese SI corpus \cite{zhao2023naist,JNPC}.
The results indicate that SI shows better word order synchronization with source language inputs than offline translation in translating relatively long sentences.
\vspace{-2mm}
\section{Related Work}
\subsection{Word Order Synchronization for Delay Reduction}
SI and SiMT should be evaluated considering both delay and accuracy.
\citet{Mieno2015SpeedOA} reported translation delay affected human evaluation results for SiMT.
One effective approach to delay reduction is synchronizing translation outputs with inputs, thereby reducing word order discrepancies between the source and target languages, as professional interpreters do \cite{he-etal-2016-interpretese,9310461,doi-etal-2021-large}.
Our work measures the word order difference to evaluate its delay.

\subsection{Evaluation of Simultaneous Interpretation}
There are two famous human evaluation metrics for SI: NAATI (National Accreditation Authority For Translators and Interpreters) Metrics \cite{naati} and EU Metrics \cite{EUmetrics}, which primarily assess the overall interpreter's performance rather than the quality of a certain interpretation result.
In contrast, this work investigates the SI word order synchronization on a sentence-by-sentence basis.

\subsection{Automatic MT Evaluation}
BLEU \cite{papineni-etal-2002-bleu} has been used in almost all previous studies in SiMT \cite{ma-etal-2019-stacl}.
RIBES \cite{isozaki-etal-2010-automatic} considers the word order difference using rank correlation coefficients.
Recent embedding-based metrics aim for semantics-oriented evaluation free from surface-level matching.
BERTscore \cite{Zhang2020BERTScore} uses token similarity considering \emph{explicit} token alignment between the hypothesis and reference.
COMET \cite{rei-etal-2020-comet} uses sentence-level embeddings
leveraging a multilingual pre-trained model, and it has been extended to reference-free COMET-QE \cite{rei-etal-2021-references}.
Our method is motivated by RIBES, BERTScore, and COMET-QE for measuring word order synchronization.

\subsection{Latency Evaluation}
Previous SiMT studies measure the latency using Average Lagging \cite{ma-etal-2019-stacl} and its variants \cite{arivazhagan-etal-2019-monotonic,papi-etal-2022-generation}.
These metrics measure the SiMT delay without considering the input and output semantic correspondence.
SI studies use Ear Voice Span \cite[EVS;][]{robbe2019ear} as an intuitive latency measure based on the time lag between the original utterances and the corresponding SIs.
This work quantifies the word order synchronization that should affect the latency in SI and SiMT for investigating the techniques to reduce the latency, instead of using timestamps.


\section{Prerequisites}
As mentioned earlier, we measure the word order synchronization between the input and output.
We use the following techniques used in previous automatic MT evaluation metrics.

\subsection{Token Alignment in BERTScore}\label{subsec:alignment}
BERTScore uses greedy token alignment for hypothesis tokens $h_i$ against reference tokens $r_1, ..., r_n$ as $a_i = \argmax_{j} \cos (\pmb{h}_i, \pmb{r}_j)$,where $a_i$ stands for the index of the reference token aligned to $h_i$, $\pmb{h}_i$ and $\pmb{r}_j$ are the embedding vectors.
We can easily apply the original monolingual BERTScore to the cross-lingual situation using multilingual BERT.

Although bilingual word alignment \cite{brown-etal-1993-mathematics, jalili-sabet-etal-2020-simalign, dou-neubig-2021-word} is popular for this problem, we did not obtain satisfactory results in our preliminary experiments
probably due to surface-level non-parallelism in SI, including summarizations, omissions, etc \cite{he-etal-2016-interpretese}.

\subsection{Word Order synchronization as Rank Correlation}\label{subsec:rank_correlation}
\citet{isozaki-etal-2010-automatic} measures the word order synchronization using Kendall's $\tau$ rank correlation coefficients between hypothesis and reference.
Once we have explicit word alignment between hypothesis and reference, we can
calculate rank correlation coefficients using the indices of aligned words.

\section{Proposed Method}
\label{section4}
\subsection{\red{Synchro Metric}}
\label{sec:one}
We propose a \red{word order synchronization metric, abbreviated as the `synchro'} for SI and SiMT based on cross-lingual token alignment, enabling us to obtain the indices of maximum similarity among all tokens although the source and target languages are different.	


Suppose we have an English input ``I ate apples yesterday.'' and the corresponding Japanese translation
\begin{CJK}{UTF8}{min}
``{私は (\textit{I}) / 昨日 (yesterday) / りんごを (apple) / 食べました。(ate)},''
\end{CJK}
where slashes represent boundaries of English words and Japanese chunks (in \emph{bunsetsu}).
Note that this is a simplified example and the actual tokenization is in subwords.
If we have the correct alignment,
we obtain the integer list [1, 4, 3, 2] to represent the word reordering happened in the translation.
In this work, we use Spearman’s rank correlation coefficient $\rho$ instead of Kendall's $\tau$ to take the reordering distance into account.
In the example, $\rho$ is equal to $0.2$, which indicates little rank correlation.

Here, we suffer from many noisy alignments due to the non-parallelism mentioned above.
To address this, we apply two heuristics to identify reliable bilingual alignment.
First, we discard alignments with English function words because they have minimal lexical significance or convey grammatical connections among other words within a sentence, which is possible for a word or token present in the source to be absent in the target, or vice versa.
We identify function words based on their part-of-speech tags using the function word list of SpaCy. 

Second, we ignore non-reliable alignments with cosine similarity below a certain threshold $\theta$.
Multilingual BERT is suitable for getting token alignment on non-English language, however due to the limitation of Multilingual BERT performance and uniqueness in SI such as summarizations and omissions, low maximum similarity scores often end up in matching token incorrectly.
To mitigate this concern, filtering with the words having high maximum similarity is necessary to accurately calculate rank correlation coefficient.


\subsection{\red{Combined Metric\label{sec:second}}}
\red{We propose combined metrics of content words coverage and word order synchronization, abbreviated as the `combined', as automatic simultaneous interpretation evaluation.
Unlike synchro, this approach distinguishes itself through alignment method enhancements and the incorporation of content word coverage }

To enhance its accuracy in calculation of rank correlation coefficients, we have improved word alignment method.
In the first step, we tokenize a sentence into words using spaCy.
Each word is then tokenized into subwords using the mBERT tokenizer.
We proceed to obtain bilingual subword alignment using both Awesome Align and BERTScore.
Subsequently, we extract subword alignment by identifying common alignments between Awesome Align and BERTScore, converting these subword alignments to word-level alignments.
Based on the word alignment, we calculate word order synchronization by calculating Spearman's rank correlation coefficient $\rho$.


\red{In addition, this metric introduces the coverage of content words.
Our initial experiments of calculating word order synchronization among different SI ranks revealed that assessing word order synchronization is essential but not enough to capture the quality of the outputs, leading to the introduction of the content words coverage as an additional metric.
We introduce content word coverage as follows to the idea from RIBES's penalty.
 Content Words Coverage $= \frac{n}{N} $
, where $n$ is the number of content words obtained in target outputs, and $N$ is the total number of content words in the source inputs.}

\begin{figure}[t]
    \centering
    \begin{adjustbox}{width=0.50\textwidth}
        \includegraphics{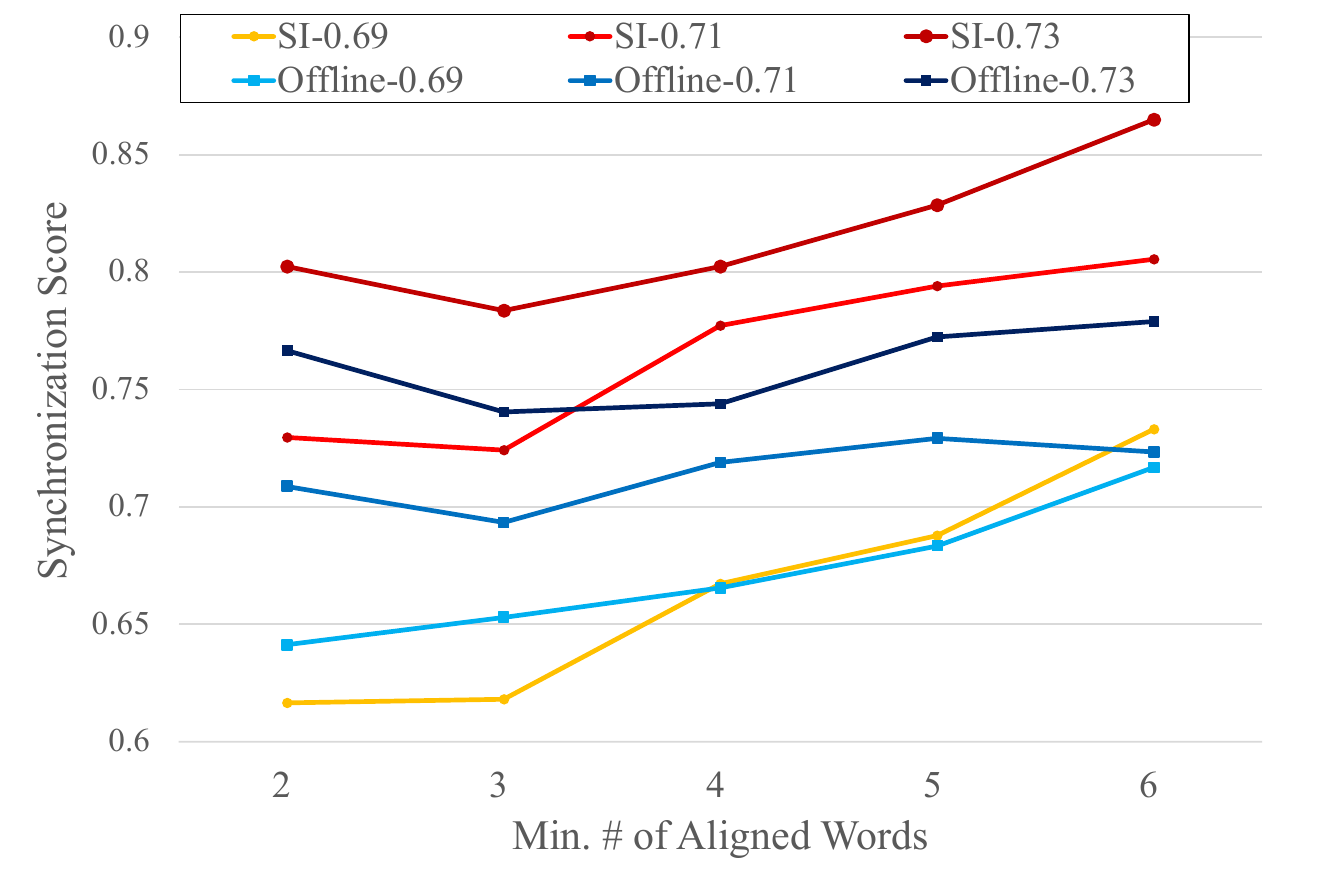}
    \end{adjustbox}
    \caption{Proposed metrics scores with varying the alignment threshold.}
    \label{fig:threshold}
\end{figure}

\section{Experiments}
\subsection{\red{Comparison between SI and Offline Translation by Synchro Metric}}

\subsubsection{Setup}
\label{sec:one_one}
We conducted the following two experiments to investigate the effectiveness of the synchro metric on actual SI and offline translation. 
In the first experiment, we compared offline translation with SI using the proposed metric.
We used NAIST-SIC-Aligned \cite{zhao2023naist}, a collection of English-to-Japanese SI for TED Talks with automatic segment-level alignment.
It consists of triples of English talk transcripts, Japanese offline translation subtitles, and transcripts of Japanese SI by professional interpreters.
We extracted 212 talks interpreted by interpreters with more than 15 years experience to focus on high-quality interpretations.
SI by less experienced interpreters made it difficult to find semantic correspondence robustly due to interpretation errors.
We reserve analyses of the less experienced interpretations as future work.

In the second experiment, we investigated the correlation of the proposed metric to human judgment.
We used one talk session from Japan National Press Club (JNPC) Interpreting Corpus \citelanguageresource{JNPCgsk}, a collection of English-to-Japanese SI for press conferences.
The SI transcripts were evaluated by a Japanese-native professional interpreter using Multidimensional Quality Metrics (MQM) \cite{lommel_2014,agrawal-etal-2023-findings} from the viewpoint of her SI expertise.
This is based on our previous work when we hired a translation evaluator and SI evaluator to evaluate SI output with MQM, we observed different viewpoints between the two regarding accuracy and fluency.
Particularly in fluency, the translation evaluator assigned negative scores to outputs if the target speech's word order synchronized with the source speech, judging it as unnatural and damaging its fluency, whereas the SI evaluator did not share the same opinion.
We consider these differences to be distinctive features specific to SI and asked evaluators to evaluate SI outputs with MQM, taking into account the unique characteristics of SI.
We chose 25 segments out of 172, which are longer than 30 words and have more than three aligned words, according to our findings in the first experiment.

\subsubsection{Analysis}

\begin{table}[t]
\centering
\begin{tabular}{ccc}
\toprule
\multirow{2}{*}{$N_{align}$} & Offline & Simultaneous \\
& translation & interpretation \\
\midrule
2 & 0.7087 (4614) & 0.7297 (2269) \\
3 & 0.6933 (2823) & 0.7242 (1014) \\
4 & 0.7188 (1598) & 0.7772 (428)\\
5 & 0.7293 (945) & 0.7941 (175) \\
6 & 0.7233 (510) & 0.8054 (86) \\
\bottomrule
\end{tabular}
\caption{\label{tab:main_result} Proposed metric scores for SI and offline translation  with $\theta{=}0.71$, varying $N_{align}$ (the minimum number of aligned English words). The numbers in parentheses indicate the number of segments evaluated.}
\end{table}

\begin{table*}[ht]
\centering
\begin{tabular}{cp{13cm}}
\toprule
Source & {\small (1) I learned / (2) new characters / (3) every day / (4) during the course of the next 15 years.} \\
\midrule
Simultaneous & \begin{CJK}{UTF8}{min}{\small (4) それから15年 (\textsl{during the course of the next 15 years}) / (3) 毎年ずっと (\textsl{every day}) / }\end{CJK} \\
Interpretation & \begin{CJK}{UTF8}{min}{\small (2) 新しい文字を (\textsl{new characters}) / (1) 学んできました。(\textsl{I learned})}\end{CJK} \\
\midrule
Offline & \begin{CJK}{UTF8}{min}{\small (4)  その後15年間 (\textsl{during the course of the next 15 years}) / (3) 毎日 (\textsl{every day}) / }\end{CJK} \\
Translation & \begin{CJK}{UTF8}{min}{\small (2) 新しい漢字を (\textsl{new characters}) / (1) 習いました。(\textsl{I learned})}\end{CJK} \\
\bottomrule
\end{tabular}
\caption{\label{tab:tau-example_short} Example for a short segment}
 \end{table*}
\vspace{5mm}

\begin{table*}[ht]
\centering
\begin{tabular}{cp{13cm}}
\toprule
Source & {\small (1) Now mathematicians / (2) have been hiding and writing / (3) messages in / (4) the genetic code / (5) for a long time / (6) but / (7) it's clear / (8) they were mathematicians and not biologists / (9) because if you write long messages with / (10) the code / (11) that the mathematicians developed / (12) it would more than likely lead to / (13) new proteins being synthesized / (14) with unknown functions.}\\
\midrule
Simultaneous & \begin{CJK}{UTF8}{min}{\small (1) 数学者は ( \textsl{Now mathematicians}) / (3) この様なメッセージを ( \textsl{messages in} ) / (4) 遺伝}\end{CJK} \\
Interpretation & \begin{CJK}{UTF8}{min}{\small 子コードで ( \textsl{the genetic code}) / (2) 作って来たんです。 ( \textsl{have been hiding and writing}) / (6) けどもしかし (\textsl{but}) / (11) 数学者は生物学者ではありません。(\textsl{Mathematicians are not biologists.}) そして間違ってる物もある訳です / (13) 新しいタンパク質を合成してしまう訳です。(\textsl{It means synthesizing a new protein.})}\end{CJK} \\
\midrule
Offline & \begin{CJK}{UTF8}{min}{\small (5) 長い間 (\textsl{for a long time}) / (4) 遺伝子コードに (\textsl{the genetic code}) / (3) メッセージを書き込む}\end{CJK} \\
Translation & \begin{CJK}{UTF8}{min}{\small 仕事は (\textsl{messages in}) / (1) 数学者が (\textsl{Now mathematicians}) / (2) 行ってきました。 (\textsl{have been writing}) / (8) 数学者は生物学者ではありません (\textsl{Mathematicians are not biologists}) / (11) 数学者が作成した (\textsl{the mathematicians developed}) / (10)コードを使って (\textsl{the code}) / (9) 長いメッセージを書いたとすると (\textsl{if you write long messages with}) / (14) 未知の機能を持った (\textsl{with unknown functions}) / (13) 新しいタンパク質の合成に (\textsl{new proteins being synthesized}) / (12) つながることでしょう。(\textsl{It would more than likely lead to})}\end{CJK} \\
\bottomrule
\end{tabular}
\caption{\label{tab:tau-example_long} Example for a long segment}
\end{table*}

\begin{table}[t]
\centering
\begin{tabular}{ccc}
\toprule
Metric & Score \\
\midrule
COMET-QE & 0.161 \\
BERTScore(F1) & 0.1111\\
Proposed & \textbf{-0.497}\\
\bottomrule
\end{tabular}
\caption{\label{tab:correlation} Pearson correlation to MQM-based human judgement with a focus on long segments}
\end{table}

For the first experiment, Figure~\ref{fig:threshold} shows the results with different alignment thresholds $\theta$, and Table~\ref{tab:main_result} shows the details with $\theta{=}0.71$.
We compared the scores for different subsets in the whole dataset varying the number of aligned English words ($N_{align}$) for two reasons.
One is that the number of elements affects the stability of the correlation coefficients.
The other is that a long sentence with large $N_{align}$ is affected by the reordering more than a short one.
Here, note that the scores with the same $N_{align}$ cannot be compared directly because of the difference in the evaluated subsets.

The results show larger $\theta$ and $N_{align}$ resulted in larger metric scores and stronger word order synchronization in SI.
These findings suggest that SI outputs synchronize the input word order more than offline translation for long sentences in which we can identify many reliable word alignments.
On the contrary, the difference between offline and SI was not so large with small $N_{align}$.

We discuss the findings above using examples.
Table~\ref{tab:tau-example_short} shows an example for a short segment.
We can see the interpretation reversed the word order same as the offline translation.
This suggests the interpreter can choose natural Japanese word order for this short input, which results in a small metric score discrepancy between SI and offline translation.
In contrast, as shown in the long segment example in Table~\ref{tab:tau-example_long}, the SI was kept aligned with the original content order in English more than the offline translation.
This is due to the difficulty in long distance reordering in SI due to the time constraints and the limitation of the interpreter's working memory, which is also suggested by some omissions.
The offline translation includes more reordering for maintaining fluency, even though this translation seems to be synchronized with the input without reordering the latter part of the segment starting with ``because'' to the beginning, probably came from the translation style in TED Talks \red{employing subtitles, which differs slightly from offline translation.}
This kind of difference should result in the metric score discrepancy observed in Table~\ref{tab:main_result}.

Finally, we investigated the relationship between word order synchronization and human judgment of the interpretation quality.
Table~\ref{tab:correlation} shows the Pearson correlation coefficients between the MQM error score and the metrics scores (COMET-QE, BERTScore (F1), and the proposed metric).
The MQM is a less is better metric, meaning it is based on error analysis. 
A lower score indicates better quality, while a higher score reflects worse quality. 
In contrast, the proposed and other metrics follow a more is better metrics, where a higher score means better quality. 
When comparing error-based scores with accuracy-based scores, a negative correlation suggests that the two scoring methods are aligned, while a positive correlation indicates they are not.
Using this interpretation, the proposed metric demonstrated a negative correlation to human judgment, while the others showed little correlation probably due to the difficulty of quality measurement of SI using the current automatic MT evaluation metrics.
This result supports the usefulness of the measurement of word order synchronization for SI evaluation.
The findings highlight the need for a specialized approach to the evaluation of SI and SiMT not in the word-by-word semantic equivalence but in the appropriate content delivery under real-time constraints.

Furthermore, we compare the efficacy of the MQM modified with SI manner to the original MQM, designed for evaluating translation outputs.
Table~\ref{tab:correlation2} presents the Pearson correlation coefficients, revealing a stronger correlation with MQM modified with SI manner.
Similar to the score interpretation in Table~\ref{tab:correlation}, Table~\ref{tab:correlation2} shows that the MQM modified with SI manner has a stronger negative correlation, suggesting that the modified version of MQM is more suitable for evaluating SI outputs than the original MQM, which was primarily designed for assessing translation quality.

\subsection{\red{Comparison among different SI ranks\label{sec:diff} by Combined Metric}}
\subsubsection{Setup}

\begin{table}[t]
\centering
\begin{tabular}{cc}
\toprule
Metric & Score\\
\midrule
MQM & \\
(original) & 0.132\\
\hline
MQM & \\
(SI modified) & \textbf{-0.497}\\
\bottomrule
\end{tabular}
\caption{\label{tab:correlation2} Pearson correlation of MQM between SI modified and original.}
\end{table}

\begin{table*}[ht]
\centering
\begin{tabular}{ccccc}
    \hline
    \multirow{1}{*}{Model} & AER & Precision & Recall & F1 \\
    \hline
    Best model ($\alpha$-entmax, multilingual ($\beta$ = 0)) & 37.4 & 72.7 & 55.0 & 0.626 \\
    Model without fine-tuned softmax & 45.5 & 68.0 & 45.5 &  0.545 \\
    \textbf{Combined} & 52.3 & \textbf{73.5} & 35.3 &  0.477 \\
    \hline
\end{tabular}
\caption{Effectiveness of Word Alignment}
\label{tab:combined}
\end{table*}

\begin{table*}[ht]
\centering
\begin{tabular}{ccccc}
    \hline
    \multirow{1}{*}{Talk Name} & A & B & S & Caption \\
    \hline
    AlexanderWagner\_2016X & 0.611 (0.433) & 0.799 (0.493) & 0.593 (0.480) & 0.465 (0.623) \\
    CharlesFleischer\_2005 & 0.574 (0.438) & 0.637 (0.4644) & 0.679 (0.455)& 0.605 (0.617) \\
    ColinGrant\_2012X & 0.543 (0.338) & 0.547 (0.470) & 0.614 (0.425) & 0.594 (0.587) \\
    DilipRatha\_2014G & 0.482 (0.488) & 0.561 (0.526) & 0.575 (0.507) & 0.553 (0.604) \\
    HamishJolly\_2013X & 0.473 (0.334) & 0.566 (0.328) & 0.526 (0.391) & 0.575 (0.664) \\
    \textbf{IlonaSzabodeCarvalho\_2014G} & 0.328 (0.373) & \textbf{0.539 (0.417)} & \textbf{0.583 (0.438)} & 0.508 (0.646) \\
    JonathanDrori\_2010U & 0.592 (0.461) & 0.541 (0.465) & 0.694 (0.485) & 0.657 (0.623) \\
    LaurelBraitman\_2014S & 0.425 (0.360) & 0.566 (0.3932) & 0.548 (0.401) & 0.586 (0.600) \\
    MiruKim\_2008P & 0.517 (0.437) & 0.521 (0.428) & 0.412 (0.470) & 0.412 (0.668) \\
    NickBostrom\_2015 & 0.487 (0.358) & 0.652 (0.375) & 0.675 (0.385) & 0.517 (0.611) \\
    ShlomoBenartzi\_2011S & 0.422 (0.433) & 0.477 (0.422) & 0.502 (0.448) & 0.605 (0.630) \\
    ShubhenduSharma\_2016S & 0.498 (0.444) & 0.663 (0.507) & 0.415 (0.487) & 0.610 (0.626) \\
    \textbf{TaraWinkler\_2016X} & 0.420 (0.444) & \textbf{0.780 (0.483)} & \textbf{0.595 (0.408)} & 0.568 (0.613) \\
    \hline
\end{tabular}
\caption{Each Talk's Word Order Synchronization (Content Words Coverage) among different SI ranks and Caption}
\label{tab:eachtalk}
\end{table*}

\begin{table*}[ht]
\centering
\begin{tabular}{cp{11cm}}
   \hline
    Source & {\small And in the case of drugs in order to undermine this fear and prejudice that surrounds the issue we managed to gather and present data that shows that today's drug policies cause much more harm than drug use per se and people are starting to get it .} \\
    \hline
    Target & {\small \begin{CJK}{UTF8}{min}ドラッグに関して恐怖であったり偏見をなくして行く為にデータを見せて行く今の政策の方がドラッグよりもどんどん 悪い結果になっているそしてそれが人々には分かり始めました。\end{CJK}}  \\
    \hline
    Aligned Words & {\small cause, harm, drugs, drug, people, fear, starting, prejudice, data, today, policies} \\
    \hline
    Synchro & {\small 0.9818}\\
    \hline
    Coverage & {\small 0.478}  \\
    \hline

\end{tabular}
\caption{\label{tab:tau-example_short_combined1} Example of the sentence achieving high synchronization and coverage}
\label{tab:detailed}
\end{table*}
\vspace{5mm}

\begin{table*}[ht]
\centering
\begin{tabular}{cccc}
    \hline
    \multirow{1}{*}{Talk Name} & SI rank & MQM (SI-modified)  & Combined \\
    \hline
    IlonaSzabodeCarvalho\_2014G & B & 238 & 0.225 \\
    IlonaSzabodeCarvalho\_2014G & S & 92 & 0.255 \\
    TaraWinkler\_2016X & B & 10 & 0.342 \\
    TaraWinkler\_2016X & S & 105 & 0.243 \\
    \hline
\end{tabular}
\caption{Comparing human evaluation to combined metric with a focus on two talks. In MQM, lower scores are better, whereas in the Combined metric, higher scores are preferable.}
\label{tab:two}
\end{table*}

We conducted the following two experiments to investigate the combined metric's effectiveness among different SI ranks. 
In the first experiment, we used the Kyoto Free Translation Task (KFTT) \cite{kftt}, a task for Japanese-English translation that focuses on Wikipedia articles related to Kyoto to investigate the effectiveness of the improved word alignment method.
Among the datasets, we extracted test data from Alignment Data.

In the second experiment, we used NAIST-SIC-Aligned \cite{zhao2023naist}, a collection of English-to-Japanese SI for TED Talks with automatic segment-level alignment to investigate a combined metric's effectiveness.
It consists of triples of English talk transcripts, Japanese offline translation subtitles, and transcripts of Japanese SI by professional interpreters.
We extracted 13 talks because these talks are only talks translated by all different experienced interpreters(S, A, and B).

The process of human judgment follows the same as in the first metrics' proposal.

\subsubsection{Analysis}
In the first experiment, Table~\ref{tab:combined} illustrates the effectiveness of the improved word alignment method.
Despite the lowest values in AER, Recall, and F1, our method shows the highest Precision.
This suggests that when the proposed model predicts a positive instance, it tends to be accurate.
However, it also indicates a notable omission of actual positive instances.
The further analysis of its word alignment is in Appendices ~\ref{sec:word_alignment}.

Our proposed method, though providing fewer alignments, maintains a relatively high accuracy, making it suitable for this task.
Therefore, we use this word alignment method for the rest of the research.

In the second experiment, Table~\ref{tab:eachtalk} presents the word order synchronization and content words coverage for each talk across various SI ranks and Captions.
For word order synchronization, there is a relative synchronization between source inputs and target outputs in SIs, better than in Caption.
As for the content words coverage, Caption shows the highest coverage, which is an anticipated finding.
However, what is noteworthy is that in certain talks, the coverage is higher in the B rank than in the S rank.
This observation is surprising, as we initially hypothesized that higher ranks would correspond to better quality.

To further investigate this, we provide a detailed analysis using an example shown in Table~\ref{tab:detailed}.
Our examination resulting from this experiment reveals that regardless of the rank of the SI, if a sentence achieves both high synchronicity and coverage, the outputs tend to demonstrate high quality.

Finally, we examine whether the results obtained from our second proposal correlate with human judgment.
{In this analysis, we chose two talks IIlonaSzabodeCarvalho\_2014G and TaraWinkler\_2016X with a focus on S rank and B rank interpreters' outputs.
This choice was based on the observation that, contrary to our initial intuition, in one talk, S rank interpreters performed better, while in the other, B rank interpreters exhibited superior output.

In Table~\ref{tab:eachtalk}, the combined metric in IIlonaSzabodeCarvalho\_2014G reveals that the S rank outperforms the B rank.
Specifically, the synchronization score and content word coverage for the S rank are 0.583 and 0.438, respectively, while those for the B rank are 0.539 and 0.417.
In terms of human evaluation scores, the same tendency is observed in Table~\ref{tab:two}, with penalty scores for S rank interpreters at 92, as opposed to 238 for B rank interpreters. 
Consequently, both quality evaluation reveals that the output of S rank interpreters is superior to that of B rank interpreters. 

On the other hand in TaraWinkler\_2016X combined metric indicates that B rank surpasses S rank.
Notably, the B rank achieves a synchronization score of 0.780 and a content words coverage of 0.438, while the S rank scores 0.595 and 0.408 in the same metrics.
Human evaluation scores represent the same trend, with penalty scores of 10 for B rank and 105 for S rank interpreters, showing the superiority of B rank outputs.
This result indicates that, despite having the most experience, the S rank doesn't consistently deliver the best performance.

Therefore, these observations suggest the importance of prioritizing the investigation of output quality over years of experience for a fair evaluation.
Detailed analyses with examples are in Appendix ~\ref{sec:detailed_analysis}.
Additional investigation into the word alignment method is also provided in the  Appendix ~\ref{sec:alignment_analysis}}.


\section{Conclusion}
In this work, we proposed a word order synchronization metric, abbreviated as the ’Synchro metric', employing BERTScore to obtain token alignments directly from the source and target.
We also introduced combined metrics of content words coverage and word order synchronization, abbreviared as combined metric, leveraged alignment information from Awesome Align and combined it with BERTScore to obtain reliable token alignments.

Our first metric revealed stronger word order synchronization of SI for long input segments than that of offline translation through the experiments using an English-to-Japanese SI corpus.

Our second metric demonstrated that outputs achieving both high word order synchronization and high coverage of content words are a sign of high-quality outputs in some cases.
It also revealed high word order synchronization and high content words coverage are not enough for the high quality outputs in some cases when the significant information is omitted and hurts overall understanding.

Word order synchronization between the source and target has not received sufficient attention in both SI and SiMT, despite its crucial role solving the trade-off between quality and latency.
Our focus on these issues led to the proposed token and word semantic-based evaluation, recognizing that word order gaps significantly impact latency in both SI and SiMT.
Future work includes more robust synchronization evaluation for SI and SiMT with summarizations, paraphrasing, omissions, etc., and quality evaluation specialized in SI and SiMT.

\nocite{*}
\section{Bibliographical References}\label{sec:reference}
\bibliographystyle{lrec-coling2024-natbib}
\bibliography{lrec-coling2024-example}

\section{Language Resource References}\label{lr:ref}
\bibliographystylelanguageresource{lrec-coling2024-natbib}
\bibliographylanguageresource{languageresource}

\section{Appendices}
\subsection{Additional Word Alignment}
\label{sec:word_alignment}

\begin{table*}[ht]
\centering
\begin{tabular}{cp{11cm}}
   \hline
    Sentence Pair & {\small \begin{CJK}{UTF8}{min}東西 線 の 駅 は 駅 ごと に ステーション カラー が 制定 さ れ て い る が 、 三条 京阪 駅 の ステーション カラー は 牡丹 色 。\end{CJK}} \\
    & {\small each station on the tozai line has its own color , and the color used at sanjyo-keihan station is scarlet tinged with purple .} \\
    \hline
    \multirow{1}{*}{Human-annotated} & {\small \begin{CJK}{UTF8}{min}each:ごと, each:に, station:駅, tozai:東西 \end{CJK}} \\
    {Word Alignment}& {\small \begin{CJK}{UTF8}{min}line:線, has:制定, has:さ, has:れ, has:て, has:い, has:る, its:ステーション, its:カラー, own:ステーション, own:カラー, color:ステーション, color:カラー, ,:、, color:ステーション, color:カラー, at:の, sanjyo-keihan:三条, sanjyo-keihan:京阪, station:駅, is:は, scarlet:牡丹, scarlet:色, tinged:牡丹, tinged:色, with:牡丹, with:色, purple:牡丹, purple:色, .:。\end{CJK}} \\
    \hline
    \multirow{1}{*}{Word Alignment} & {\small \begin{CJK}{UTF8}{min}station:駅, line:線, has:は, color:カラー\end{CJK}} \\ 
    {from Combined}& {\small \begin{CJK}{UTF8}{min}and:、, color:カラー, at:の, sanjyo-keihan:三条, sanjyo-keihan:京阪, station:駅, is:は, scarlet:牡丹, purple:色\end{CJK}} \\
    & \\
    \hline
    
\end{tabular}
\caption{Effectiveness of Word Alignment}
\label{tab:combined-analysis}
\end{table*}

\begin{table*}[ht]
\centering
\begin{tabular}{cp{11cm}}
   \hline
    Source & {\small But such ardent watchfulness can lead to anxiety so much so that years later when I was investigating why so many young black men were diagnosed with schizophrenia six times more than they ought to be I was not surprised to hear the psychiatrist say Black people are schooled in paranoia.} \\
    \hline
    \multirow{1}{*}{Target} & {\small \begin{CJK}{UTF8}{min}不安感がありましたそれで何年か経って何で若い黒人が統合失調症になる可能性が六倍あると言う事が分かりました個人黒人と言うのはパラノイアだと言う風に言われてるんです。\end{CJK}} \\
    \hline
    {\textbf{Combined}}& {\small \begin{CJK}{UTF8}{min}watchfulness:感, anxiety:不安, years:年, young:若い, black:黒人, men:黒人, diagnosed:症, schizophrenia 失調, schizophrenia:症, times:倍, paranoia:パラノイア\end{CJK}} \\
    \hline
    \multirow{1}{*}{Awesome} & {\small \begin{CJK}{UTF8}{min}watchfulness:感, lead:あり, anxiety:不安, years:年, later:で\end{CJK}}\\
    {Align} & {\small \begin{CJK}{UTF8}{min}young:若い, black:黒人, men:黒人, diagnosed:症, schizophrenia:失調, schizophrenia:症, times:倍, hear:言う, Black:黒人, schooled:言わ, paranoia:パラノイア\end{CJK}} \\ 
    \hline
    {BERTScore}& {\small \begin{CJK}{UTF8}{min}ardent:感, watchfulness:感, lead:感,
anxiety:不安, anxiety:感, years:年, investigating:統合, investigating:が, young:若い, black:黒人, men:黒人, diagnosed:症, schizophrenia:失調, schizophrenia:症, times:倍, ought:が, surprised:言わ, surprised:に, hear:言わ, psychiatrist:パラノイア, psychiatrist:黒人, psychiatrist:症, Black:黒人 , people:黒人, schooled:統合, schooled:パラノイア, paranoia:パラノイア\end{CJK}} \\
    \hline
    
\end{tabular}
\caption{An Example of Word Alignment Method Comparison}
\label{tab:word_alignment_comparison}
\end{table*}

To investigate the proposed word alignment further, we examine word alignment through a specific example in Table~\ref{tab:combined-analysis}.
The word alignment results obtained correctly by the improved word alignment method involve content words such as station, line, and color.
Conversely, instances where the proposed method fails to obtain word alignments are characterized by function words such as each, own and is.
Given that the aim of this research is to investigate word order synchronization, with a focus on content words rather than function words and not specifically on enhancing word alignment, the observed outcome of the highest precision is enough for this study.

Additionally, Table~\ref{tab:word_alignment_comparison} shows an example of word alignment method comparison in a real scenario where source inputs and target outputs aren't in an equal relationship due to specific characteristics in SI such as omission, summarization etc.. We compared its effectiveness with a previous method and awesome align. The previous method, using BERTScore's greedy match, often aligns words incorrectly, leading to low accuracy.
Awesome align also faces misalignment issues, which can negatively impact word order distance calculations, which is the focus of our research.

\subsection{Detailed Analyses with Combined Metric}
\label{sec:detailed_analysis}

\begin{table*}[ht]
\centering
\begin{tabular}{ccc}
    \hline
    \multirow{1}{*}{Segment Length} & Combined & BERTScore (F1) \\
    \hline
    All & -0.130 & -0.049 \\
    $<15$ & 0.094 & 0.011 \\
    $\geq 15$ & -0.183 & 0.035 \\
    $\geq 20$ & -0.225 & 0.113 \\
    $\geq 25$ & -0.101 & 0.077 \\
    $\geq 30$ & -0.262 & 0.105 \\
    \hline
\end{tabular}
\caption{Rank correlation with human judgment across varying segment lengths.}
\label{tab:two_human}
\end{table*}

For a detailed analysis, we examined the rank correlation to human judgment between the proposed method and BERTScore (F1) in each sentence and the Table~\ref{tab:two_human} shows its details.

In the row 'All' in Table~\ref{tab:two_human}, a very weak correlation is observed more in combined metric than in BERTScore.
From the findings in previous results and analysis, which highlights the significant advantage of the combined metric as the segment length increased, we further breakdown its results based on varying segment lengths.
In the proposed method, we observed a slight correlation improvement as the segment length increased, whereas in BERTScore (F1), no correlation was observed regardless of the segment length.
These results indicate that our combined metric excels at evaluating relatively long sentences, though some improvement is needed when assessing short segments.

\begin{table*}[ht]
\centering
\begin{tabular}{cp{11cm}}
   \hline
    Source & {\small Each and every one of us has the power to change the world.} \\
    \hline
    Target & {\small \begin{CJK}{UTF8}{min}私達一人一人は、世界を変える力を持っているのです。\end{CJK}}  \\
    \hline
    Combined  & {\small -0.666 (-1 (Synchro) * 0.666 (Coverage))}\\
    \hline
    MQM & {\small 0}  \\
    \hline

\end{tabular}
\caption{\label{tab:tau-example_short_combined2} An example in a short segment where proposed method doesn't correlates with human judgment.}
\label{tab:detailed_short}
\end{table*}
\vspace{5mm}

Table~\ref{tab:detailed_short} shows the challenge combined metric faces when evaluating short segments, which rooted in the interaction between the two components: the multiplication of content word coverage and the word order synchronization score.
Despite a higher coverage of words, a combined metric evaluates short segments unfairly due to the unavoidable challenge of low word order synchronization in short segments, where word reordering occurs naturally even in SI. 
In long segments, a combined metric aligns with human evaluation results in some cases, while in others, it does not.

\begin{table*}[ht]
\centering
\begin{tabular}{cp{11cm}}
   \hline
    Source & {\small What I 'm talking about and what the Global Commission advocates for is creating a highly regulated market where different drugs would have different degrees of regulation.} \\
    \hline
    Target & {\small \begin{CJK}{UTF8}{min}私が言いたいと思ってるのは非常に規制の高いマーケットを作ると言う事ですドラッグのレベルによって規制のレベルを変えると言う事です。\end{CJK}}  \\
    \hline
    Combined & {\small 0.417 (0.893 (Synchro)) * 0.467 (Coverage)}\\
    \hline
    MQM & {\small 0}  \\
    \hline

\end{tabular}
\caption{\label{tab:detailed_long} An positive example in a long segment where proposed method correlates with human judgment.}
\label{tab:positive}
\end{table*}
\vspace{5mm}

\begin{table*}[ht]
\centering
\begin{tabular}{cp{11cm}}
   \hline
    Source & {\small Civil society diplomats do three things They voice the concerns of the people are not pinned down by national interests and influence change through citizen networks not only state ones.} \\
    \hline
    Target & {\small \begin{CJK}{UTF8}{min}人々の声を取り上げ国の声では無くて人々の声を取り上げて来るのです。\end{CJK}}  \\
    \hline
    Combined & {\small -0.09 (-0.5 (Synchro) * 0.188 (Coverage))}\\
    \hline
    MQM & {\small 10}  \\
    \hline

\end{tabular}
\caption{An negative example in a long segment where combined metric correlates with human judgment.}
\label{tab:detailed_omission2}
\end{table*}
\vspace{5mm}

The following is the example in the long segment when a combined metric aligns with human judgment.
In Table~\ref{tab:detailed_long}, the sentence is not penalized in human evaluation, indicating the quality of its output is high. In a combined metric, the multiplication of the average synchronization score and content words coverage, resulting in 0.255, indicating that the multiplied score of 0.417 expresses higher quality compared to the average score.
In Table~\ref{tab:detailed_omission2}, the evaluator penalizes this sentence as 10, and combined metric gives -0.009).
In such instances where obvious omission happens, the combined metric could capture its tendency due to the calculation of its content words.

\begin{table*}[ht]
\centering
\begin{tabular}{cp{11cm}}
   \hline
    Source & {\small In just a few years, we not only changed national legislation that made it much more difficult for civilians to buy a gun, but we collected and destroyed almost half a million weapons.} \\
    \hline
    Target & {\small \begin{CJK}{UTF8}{min}この何年の内にもっともっと市民が銃を買うのを難しくないといけない。そして半分の銃は捨て去って行かないといけない。\end{CJK}}  \\
    \hline
    Combined  & {\small 0.612 (0.883 (Synchro) * 0.692 (Coverage))}\\
    \hline
    MQM & {\small 10}  \\
    \hline

\end{tabular}
\caption{An example in a long segment where proposed method does not correlates with human judgment due to its mistranslation.}
\label{tab:detailed_overlook}
\end{table*}
\vspace{5mm}

\red{Table~\ref{tab:detailed_overlook} shows a examples when combined metric cannot fairly evaluate long segments.
In the example, the evaluator penalizes this sentence as 10, while combined metric gives 0.612.}
In such instances, the translations may initially seem coherent due to an adequate number of target outputs corresponding to the source inputs.
However, more detailed examination reveals the lack of connection between individual words results in outputs that are challenging to understand.
A combined metric cannot follow the connectivity of each word since we only count the content words coverage and the word order synchronization, leading to discrepancy between the human judgment and proposed method.

\begin{table*}[ht]
\centering
\begin{tabular}{cp{11cm}}
   \hline
    Source & {\small Civil society diplomats do three things They voice the concerns of the people are not pinned down by national interests and influence change through citizen networks not only state ones.} \\
    \hline
    Target & {\small \begin{CJK}{UTF8}{min}この外交官の仕事は、三つがあります。一つは、まず人の声を代弁する。そして、人に対して影響を与えていくと言う事です。\end{CJK}}  \\
    \hline
    Combined & {\small 0.353 (0.616 (Synchro)) * 0.375 (Coverage))}\\
    \hline
    MQM & {\small 10}  \\
    \hline

\end{tabular}
\caption{An example in a long segment where proposed method does not correlates with human judgment due to its omission.}
\label{tab:detailed_omission}
\end{table*}
\vspace{5mm}

{In Table~\ref{tab:detailed_omission}, evaluator penalizes this sentence as 10, while combined metric gives 0.353.
Critical omission is also difficult to detect in such cases when an adequate number of target outputs corresponding to the source inputs is observed at first glance.
This drawback could potentially be solved by assigning weights based on the importance of content words, instead of treating all words equally in the current proposed method.
While we have demonstrated a significant advantage of combined metric over BERTScore, there is room for further improvement in both short and long segments.
We reserve further enhancement of our automatic evaluation for more precise results as future work.

\subsection{{Comparison between SI and Offline Translation using word alignment method in \ref{sec:second}}}
\label{sec:alignment_analysis}

Additionally, we investigated the validity of method \ref{sec:diff} under the assumption that alignment method in \ref{sec:diff} outperforms \ref{sec:one}.
Given the previous findings that our method excels at evaluating long segments, we chose the same dataset in the second experiment in \ref{sec:one_one} because each segment in JNPC is longer than TED Talks.

\begin{table*}[ht]
\centering
\begin{tabular}{ccc}
    \hline
    \multirow{1}{*}{Segment} Length & SI & Translation \\
    \hline
    All & 0.686 & 0.455 \\
    $<15$ & 0.479 & 0.011 \\
    $\geq 15$ & 0.736 & 0.512 \\
    $\geq 20$ & 0.786 & 0.563 \\
    $\geq 25$ & 0.790 & 0.596 \\
    $\geq 30$ & 0.885 & 0.740 \\
    \hline
\end{tabular}
\caption{Word order synchronization comparison between SI and Translation.}
\label{tab:si}
\end{table*}

We conducted two experiments following the process in \ref{sec:one_one}.
In the first experiment, we compared the word order synchronization using combined metric's word alignment. 
The results are presented in Table~\ref{tab:si}. 
As expected, across the entire dataset, SI synchronizes more closely with the source speech than translation, and this tendency becomes more evident as segments get longer.

\begin{table*}[t]
\centering
\begin{tabular}{ccc}
    \hline
    Metric & Score \\
    \hline
    COMET-QE & 0.161 \\
    BERTScore (F1) & 0.1111\\
    Synchro & \textbf{-0.497}\\
    Combined (synchro only) & \textbf{-0.011}\\
    Combined & \textbf{0.022}\\
    \hline
\end{tabular}
\caption{\label{tab:si-2} Pearson correlation to MQM-based human judgement with a focus on long segments}
\end{table*}

In the second experiment, we compared synchro score alone and combined metric with human evaluation Table~\ref{tab:si-2}.
However, neither the synchro metric nor the combined metric showed any correlation with human judgment.

\begin{table*}[ht]
\centering
\begin{tabular}{|c|p{11cm}|}
   \hline
    Source & {\small And if we are to have a union that would be stable and that would be at peace because all the peoples accept that it is a union that will guarantee their security then we will have to amend the constitution and we have to work towards it.} \\
    \hline
    Target & {\small \begin{CJK}{UTF8}{min}我々は作りたい国が安定してそしてすべての国民が彼らの違う安全を保障してくれると信じてくれればそれならば憲法を改正できると思っております。\end{CJK}}  \\
    \hline
    Combined & {\small 0.573 (0.9 (Synchro}) * 0.636363636 (Coverage))\\
    \hline
    MQM & {\small 79.8}  \\
    \hline
    Comments by  & {\small \begin{CJK}{UTF8}{min}「安定して平和な連邦国家を築くのであれば」が大部分抜け, \end{CJK}}\\
    SI Evaluator & {\small \begin{CJK}{UTF8}{min}(誤訳)→「すべての国民が自分達の安全を保障してくれるのは連邦国家であると受け入れるのであれば」,\end{CJK}}\\
                & {\small \begin{CJK}{UTF8}{min}(誤訳)→「憲法を改正しなくてはならず、そのために努力しなければならない」の意\end{CJK}}\\
    \hline
\end{tabular}
\caption{An example in SI that the proposed method does not correlate with human judgment due to its omission and mistranslation.}
\label{tab:d}
\end{table*}

\vspace{5mm}

This results might follows the same tendency observed in Table~\ref{tab:detailed_overlook} and Table~\ref{tab:detailed_omission} indicating that critical mistranslation and omission are difficult to detect when an adequate number of target outputs corresponding to the source inputs is observed at first glance (Table~\ref{tab:d}). 
This highlights a current limitation in a combined metric, which treats each content word equally.
However, the positive outcome is that, owing to the improved alignment ability in combined metric, we successfully extracted 48 sentences meeting the criteria of segment lengths longer than 30 words and containing more than three aligned words. This represents an improvement compared to the synchro metric, which could only extract 25 sentences under the same conditions.


\end{document}